\documentclass[runningheads]{llncs}
\usepackage[T1]{fontenc}
\makeatletter
\newcommand{\printfnsymbol}[1]{%
  \textsuperscript{\@fnsymbol{#1}}%
}
\makeatother

\usepackage{graphicx}
\usepackage{amsmath}
\usepackage{amsfonts}
\usepackage{algorithm}
\usepackage{algpseudocode}
\usepackage{multirow}
\usepackage{subcaption}
\usepackage{booktabs}
\usepackage{multirow}
\usepackage[square,numbers]{natbib}

\begin{document}
\title{Entity Augmentation for Efficient Classification of Vertically Partitioned Data with Limited Overlap}
\titlerunning{Entity Augmentation}
%
\author{Avi Amalanshu\thanks{Equal contribution.}$^{,}$\thanks{Corresponding author. Email: \texttt{avi.amalanshu@kgpian.iitkgp.ac.in}}
\and
Viswesh Nagaswamy\printfnsymbol{1}
\and
G.V.S.S. Prudhvi\printfnsymbol{1}
\and
Yash Sirvi\printfnsymbol{1}
\and
Debashish Chakravarty
}
\authorrunning{A. Amalanshu et al.}
%
\institute{Autonomous Ground Vehicle Research Group\\Indian Institute of Technology Kharagpur\\ Kharagpur, WB 721302, India}
\maketitle              
\begin{abstract}
Vertical Federated Learning (VFL) is a machine learning paradigm for learning from vertically partitioned data (i.e. features for each input are distributed across multiple ``guest" clients and an aggregating ``host" server owns labels) without communicating raw data. Traditionally, VFL involves an ``entity resolution" phase where the host identifies and serializes the unique entities known to all guests. This is followed by private set intersection to find common entities, and an ``entity alignment" step to ensure all guests are always processing the same entity's data. However, using only data of entities from the intersection means guests discard potentially useful data. Besides, the effect on privacy is dubious and these operations are computationally expensive. We propose a novel approach that eliminates the need for set intersection and entity alignment in categorical tasks. Our Entity Augmentation technique generates meaningful labels for activations sent to the host, regardless of their originating entity, enabling efficient VFL without explicit entity alignment. With limited overlap between training data, this approach performs substantially better (e.g. with 5\% overlap, 48.1\% vs 69.48\% test accuracy on CIFAR-10). In fact, thanks to the regularizing effect, our model performs marginally better even with 100\% overlap.

\keywords{Federated Learning  \and Vertical Federated Learning \and Sample Efficiency.}
\end{abstract}
\section{Introduction}
\label{sec:intro}

Federated Learning (FL) \cite{pmlr-v54-mcmahan17a} is a recent distributed machine learning strategy. FL aims to achieve communication efficiency and data privacy by never communicating the raw data. In FL, data-owning participants (``guests") train models on their local data, coordinated and aggregated by a label-owning ``host". FL typically implies a ``horizontal" distribution, where a participant holds its own set of samples within a global dataset. Vertical Federated Learning (VFL) is a variant where parties holding different \emph{features} of the same samples collaborate without pooling data to learn joint representations. This is essential for sensitive cross-institution collaborations, such as in healthcare, emphasizing the importance of aligning records to the same entities for cohesive, privacy-preserving model training.

VFL effectively splits the parameters of a global model across the network. The host has the deeper layers and makes a prediction at each training/inference iteration. For the prediction to be meaningful, all guests must have passed their features of the same entity. But, this means they must discard data on entities not known to all participants-- potentially valuable for training local models. In systems with a small intersection, there may be insufficient samples to train a VFL model effectively, hindering VFL's scalability.

For example, cameras and traffic sensors at an intersection may struggle to detect crashes if the number of frames where the crash is visible to all cameras is small. The entity alignment process introduces significant computational overhead, hampering real-world VFL deployment at scale, affecting overall efficiency. Other challenges include data skew, where data distribution across entities varies drastically, and privacy risks during alignment despite VFL's principle of avoiding direct data sharing. This raises the question: are PSI and entity alignment truly necessary during training?

\begin{figure}
    \centering
    \includegraphics[width=0.73\linewidth]{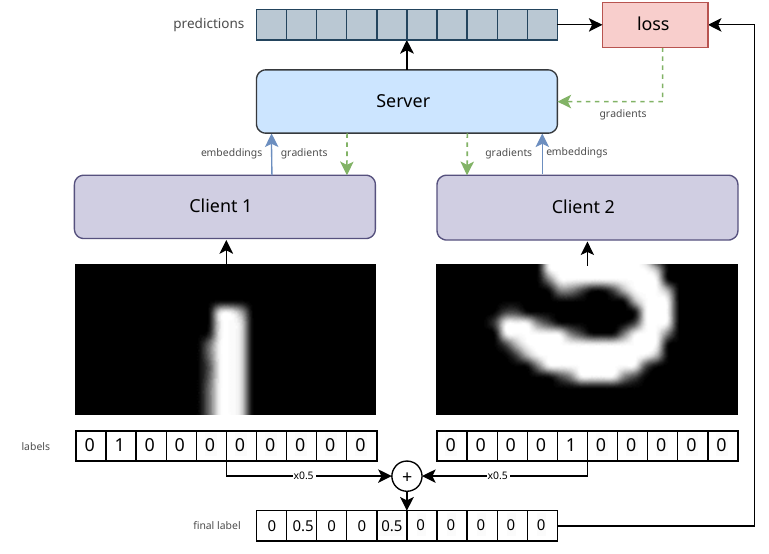}
    \caption{Example forward pass with entity augmentation. Both clients forward activations from arbitrary inputs to the host, which is aware of the identity of said inputs. Half the features in the host input correspond to the number 1 and the other half correspond to 0. The interpolated label is their weighted average.}
    \label{drawio}
\end{figure}

We introduce \emph{Entity Augmentation}, a strategy for VFL that eliminates the need for PSI and entity alignment. Instead of agreeing on a single entity (or batch), the host computes a weighted average of labels for all entities processed by any guest. The weights are proportional to the total dimension of the input vector corresponding to each entity's features. 
Hosts may calculate meaningful losses for any activations received, as long as each corresponds to labelled entities.\\
\\
In this paper, we:
\begin{itemize}
    \item Propose Entity Augmentation, a novel strategy that interpolates labels for all entities sent by all guests, weighted by their contribution to the host input, synthesizing semantically coherent labels for guest activations.
    \item Demonstrate that VFL with Entity Augmentation achieves performance on par (better on some datasets) with VFL with entity alignment.
\end{itemize}
These empirical results indicate that Entity Augmentation is a viable alternative to traditional FL pipelines, offering substantial improvements in data utilization, computational efficiency, and ease of deployment.

\section{Background}
\label{sec:background}

\subsection{VFL Participants}
\paragraph{Guests.}
Consider a consortium $\mathcal{G}$, comprising participants each with a distinct feature set. For a guest $i \in \mathcal{G}$, the dataset is $\mathcal D_i = \{\mathbf x_j \in \mathbb{R}^{|F_i|}: j \in \{1, 2, ..., |\mathcal S_i|\}\}$, where:
\begin{itemize}
    \item $\mathcal S_i$ is the set of unique entities recorded in $\mathcal D_i$.
    \item $F_i$ captures the attributes of these entities observed by guest $i$.
    \item Entities are considered samples from a distribution $X$.
\end{itemize}
The \textbf{guest model} $m_i(\cdot ; \theta_i) : \mathbb{R}^{|F_i|} \rightarrow \mathbb{R}^{\text{out}_i}$ is defined by parameters $\theta_i$.

These models aim to encode the features $F_i$ of entities $\mathbf x \in \bigcap_{i=1}^{|\mathcal{G}|} \mathcal S_i$ for the host $h$ to utilize in predictions, without sharing their model parameters or direct data features, including labels.

\paragraph{Host.} The host $h$ coordinates the training process, holding the label set $\mathcal{L} = \{\mathbf y_j \in \mathbb{R}^{\text{out}}: j \in \{1, 2, ..., |\mathcal S_h|\}\}$, where $\mathcal S_h$ is the set of unique entities with labels. A crucial intersection $|\mathcal S_{\mathcal{G}} \cap \mathcal S_h| > 0$ ensures shared entities for training.

The host model $m_h(\cdot ; \theta_h) : \mathbb{R}^{\text{out}_1} \times \mathbb R^{\text{out}_2} \times ... \times \mathbb R^{\text{out}_{|\mathcal{G}|}} \rightarrow \mathbb{R}^{\text{out}}$ is parameterized by $\theta_h$, aiming to minimize expected loss for optimal parameters $\theta = (\theta_1, \theta_2, ..., \theta_{|\mathcal{G}|}, \theta_h)$. 

\subsection{Entity Alignment}
In VFL, coherence during training is ensured through data synchronization, formalized as $\mathcal{S}_{\mathcal{G}} = \bigcap_{i=1}^{|\mathcal{G}|} \mathcal{S}_i$. This uses a private set intersection (PSI, \cite{MORALES2023100567, Lu2020MultiPartyPS}) multiparty computation, preserving privacy while identifying intersecting entities across $\mathcal{G}$.

Following PSI, the host $h$ processes $\mathcal{S}_{\mathcal{G}}$, ensuring uniform model training across the federated network. This step is vital for coherent aggregation of model updates, reflecting the collective knowledge of $\mathcal{G}$.

Without proper alignment, i.e., if $\mathcal{S}_{\mathcal{G}}$ is not established, issues like data inconsistency ($\mathcal{S}_i \not\subseteq \mathcal{S}_{\mathcal{G}}$ for any $i$) arise, leading to degraded model performance from training on non-corresponding entities. Additionally, without alignment, the federated model faces privacy vulnerabilities and inefficiencies in learning. Thus, Entity Alignment is crucial in vertical federated learning.

\section{Related Work}
\subsection{Entity Resolution in Federated Learning}

In the absence of unique IDs, the task of resolving common entities between datasets based on their features is called Entity Resolution. In 2017, Hardy et al. \cite{hardy2017private} introduced one of the first privacy-preserving strategies for learning from vertically partitioned data. The work proposes a pipeline of entity resolution, distributed logistic regression, and Paillier encryption to maintain privacy without noise addition. The authors demonstrate this works under certain entity resolution error assumptions without impacting model performance. This suggests certain errors do not alter optimal classifier performance.

Nock et al. \cite{nock2018entity} investigate the the empirical impact of entity resolution errors on FL. The authors provide bounds on deviations in classifier performance due to these errors, and demonstrate the benefits of using label information with entity resolution algorithms.

\subsection{Data Augmentation for Classification Generalization}
CutMix \cite{yun2019cutmix} is a data augmentation technique used in image classification to enhance deep learning model training by combining parts of different images and their corresponding labels. Unlike traditional methods that process each image individually, CutMix creates new training examples by patching segments from multiple images together.

Given two images \( A \) and \( B \), and their corresponding one-hot encoded labels \( \mathbf{y}_A \) and \( \mathbf{y}_B \), the CutMix process involves:

\begin{enumerate}
    \item Randomly selecting a region \( R \) within image \( A \).
    \item Replacing region \( R \) in image \( A \) with the corresponding region from image \( B \) to generate a new training image \( A' \).
    \item Combining the labels proportionally to the number of pixels of each class present in the new image, resulting in a mixed label \( \mathbf{y}' = \lambda \mathbf{y}_A + (1 - \lambda) \mathbf{y}_B \), where \( \lambda \) is the ratio of the remaining area of image \( A \) to the area of the original image.
\end{enumerate}

Mathematically, for a region \( R \) with bounding box coordinates \( (r_x, r_y, r_w, r_h) \), the new training image \( A' \) is represented as:
\begin{equation}
    A' = \begin{cases} 
    B_{r_x:r_x+r_w, r_y:r_y+r_h} & \text{for } (i, j) \in R \\
    A_{i, j} & \text{otherwise}
    \end{cases}
\end{equation}

Here, \( (i, j) \) is the pixel location in the images. The label mixing coefficient \( \lambda \) is typically sampled from a Beta distribution, which controls the strength of the mixing.

CutMix improves model robustness and generalization by forcing the network to learn regionally informative features, rather than relying on specific patterns in the training set. This generates diverse examples within each mini-batch, helping to prevent overfitting.


\subsection{Sample Efficient Vertical Federated Learning}
Work on sample efficiency is scarce, despite its absence greatly limiting the applicability of VFL to carefully designed systems with significant overlap in sample spaces. 

Sun et al. propose a method \cite{Sun_2023_ICCV} to solve this problem. Following a few epochs of VFL training on aligned data, guests cluster their remaining datasets based on gradients received during the aligned training. The authors experimentally show that this approach is performant. However, as suggested by Amalanshu et al. \cite{amalanshu2024decoupled} this is a form of privacy-breaching label inference attack. 

In that paper, the authors present an unsupervised method of training guest models independently from host models, hence allowing them to exploit data outside the intersection without breaching privacy. However, task-relevant transfer learning still uses aligned datasets.

\section{Proposed Method}
VFL typically assumes that the input datasets for each model are ``aligned," meaning that records are consistent across entities indexed in $\left(\bigcap_{i=1}^{|\mathcal{G}|}\mathcal S_{i}\right)\cap \mathcal S_{h}$. We propose a novel training approach for categorical tasks that allows each dataset to be sized $\min_{i\in\{1,\ldots,|\mathcal{G}|\}}|\mathcal S_{i}\cap \mathcal S_{h}|$, or $\max_{i\in\{1,\ldots,|\mathcal{G}|\}}|\mathcal S_{i}\cap \mathcal S_{h}|$ if guests may reuse data.
    
Extending the idea of the CutMix regularization, we propose entity augmentation for training the owner model. We construct artificial entity samples by combining features from various entities and averaging their labels. This approach enables training on a minimal subset of samples. 

There are various ways such a scheme might be implemented. For instance, entity augmentation may be precomputed before training begins-- the host may inform the guests which order to process their entities, and memoize the corresponding augmented labels. Alternatively, the augmented labels could be computed at training time as long as the host is aware of the identities of all the entities whose encoded features it has just received. Algorithm \ref{alg:EntityAug} outlines one way of achieving the latter for models trained via gradient-based algorithms. 

Using a queue to store the latest activations and sample IDs, we also achieve some fault tolerance-- if a guest fails to send an activation, the host simply uses the last one received. We outline the procedure for entity alignment and augmentation in categorical tasks. 

The proposed method optimizes data use, enhancing the robustness and generalization of the learned models. Empirical results demonstrating the effectiveness of our approach, including in scenarios with deliberate sample misalignment, are presented in Section \ref{sec:experiments}.

\begin{algorithm}[]
       \caption{Neural Network Training with Entity Augmentation}
       \label{alg:EntityAug}
       \begin{algorithmic}[1]
          \Require{$\mathcal D_i\forall i \in \{1,2,\dots,|\mathcal G|, h\}$}: Datasets of guests $i\in\mathcal{G}$ and host $h$
          \Require{$\mathcal S_{i}\ \forall i \in \{1, 2, \dots, |\mathcal{G}|\}$: Serialized sets of of entities for which guests $i$ have features}
          \Require{$\mathcal S_{h}$: Set of entities for which the label owner has labels}
          \Require{Label set $\mathcal L = \{\mathbf y_j \in \mathbb{R}^{c}:  \mathbf y_j$ is the one-hot label for entity $\mathbf x_j\}$}
          \Require{$\text{optim}_i\forall i\in  \{1,2,\dots,|\mathcal G|, h\}$}: parameter optimizer for each participant
        \end{algorithmic}
        \textbf{Guest training iteration} (for guest $i$)
        \begin{algorithmic}[1]
          
          \State Retrieve the features $\mathbf x_{j,i}$ of the next entity $\mathbf x_j$ in its dataset
          \State Calculate guest model output $\mathbf a_i\gets m_i(\mathbf x_{j,i}; \theta_i)$
          \State Send $\mathbf a_i$ and sample ID $j$ to host
          \State Receive loss gradient $\nabla_{\mathbf a_i}\ell$ from the host
          \State Perform backpropagation to obtain $\nabla_{\theta_i}\ell$
          \State Calculate weight update $\theta_i\gets\text{optim}_i\left(\nabla_{\theta_i}\ell, \theta_i\right)$
        \end{algorithmic}
        \textbf{Server executes}
        \begin{algorithmic}[1]
        \State Initialize empty activation queues $Q_i$ and label queues $Q_{\text{label}, i}$ for each guest.
        \Repeat
            \For{all guests $i\in \mathcal G$ in parallel}
                \State{Initiate guest training iteration} \Comment{send $\mathbf a_i$, $j$}
                \State {Add $\mathbf a_i$ to $Q_i$ and $j$ to $Q_{\text{label}, i}$}
            \EndFor
            \State {Read $\hat{\mathbf a}_i\forall i$ from the top of each $Q_i$}
            \State {Calculate prediction $\mathbf y\gets m_h(\hat{\mathbf a}_1,\dots,\hat{\mathbf a}_{|\mathcal G|}; \theta_h)$}
            \State {Read $\mathbf j_i \forall i$ from the top of each $Q_{\text {label}_i}$}
            \State {Retrieve label $\mathbf y_{j_i}\forall j_i$ read}
            \State {Form label $\mathbf y = \frac{\displaystyle\sum_{i=1}^{|\mathcal G|}w_i\mathbf y_{j_i}}{\displaystyle\sum_{i=1}^{|\mathcal G|}w_i}$ where $w_i$ is the dimension of $\mathbf a_i$} \Comment{Entity Augmentation}
            \State {Compute loss $\ell$}
            \State {Perform backpropagation to obtain $\nabla_{\theta_h}\ell$ and $\nabla_{\mathbf a_i}\ell\forall i\in\{1,2,\dots,|\mathcal G|\}$}
            \State {Send all gradients to their respective participants.}
            \State {Calculate weight update $\theta_h\gets\text{optim}_h\left(\nabla_{\theta_h}\ell, \theta_h\right)$}
            \For{all guests $i\in \mathcal G$ in parallel}
                \State{Complete guest training iteration}
            \EndFor
        \Until{convergence or a fixed number of iterations}
       \end{algorithmic}
    \end{algorithm}


\section{Experiments}
\label{sec:experiments}

To evaluate the effectiveness of the proposed algorithm, we conduct experiments on six different real-world datasets using three distinct architecture models in a SplitNN fashion. \cite{Gupta2018, ceballos2020splitnndriven} The experiments are divided into the following setups: (1) aligned data setup, where the dataset is entity-aligned; and (2) misaligned data setup, where the dataset is entity-augmented/misaligned. This division helps us mimic real-world scenarios where data may not always be perfectly aligned between clients.\\
\\
We hope to demonstrate the following:
\begin{enumerate}
    \item Entity Augmentation leads to meaningful learning, that is, Entity Augmentation allows us to exploit data outside the intersection $\mathcal S_\mathcal G\cap S_{h}$ (namely, members of $\bigcup_{i=1}^{|\mathcal G|} \left(\mathcal S_i\cap\mathcal S_h\right)$).
    \item Training on datasets with Entity Augmentation and without alignment outperform that on aligned datasets if there are sufficiently long-range semantic correlations.
\end{enumerate}
We also provide a brief comparison to few-shot VFL \cite{Sun_2023_ICCV} in Table \ref{tab:fewshot}.

\subsection{Datasets}
We use the following datasets and architectures for our experiments:
\begin{itemize}
    \item \textbf{Computer Vision (CV) Datasets}: MNIST \cite{lecun2010mnist} and CIFAR-10 split into two guests. \cite{Krizhevsky09learningmultiple} with ResNet-18, ResNet-56 \cite{he2015deep}, and ResNeXt-29 (8x64d) \cite{xie2017aggregated}.
    \item \textbf{Tabular Datasets}: Parkinsons \cite{sakar2019comparative} and Credit Card \cite{yeh2009comparisons}.
    \item \textbf{Multiview Datasets}: Handwritten Digits \cite{Dua:2019} and Caltech-7 \cite{li_andreeto_ranzato_perona_2022}.
\end{itemize}
The tabular and multiview datasets are divided evenly across four guests. 
 
\subsection{Model Details}
\textbf{{Models used for VFL datasets}}\\
\textbf{Handwritten.}
Guests: \textsf{linear}(120) $\to$ \textsf{linear}(70) $\to$ \textsf{ReLU};
Hosts: \textsf{linear}(280) $\to$ \textsf{linear}(120) $\to$ \textsf{LeakyReLU}$\to$ \textsf{linear}(40) $\to$ \textsf{linear}(10)\\
\textbf{CalTech-7.}
Guests: \textsf{linear}(512) $\to$ \textsf{linear}(256) $\to$ \textsf{ReLU};
Hosts: \textsf{linear}(1024) $\to$ \textsf{linear}(512) $\to$ \textsf{linear}(256) $\to$ \textsf{LeakyReLU} $\to$ \textsf{linear}(128) $\to$ \textsf{linear}(7)\\
\textbf{Credit Card.}
Guests: \textsf{linear}(5) $\to$ \textsf{linear}(2) $\to$  \textsf{ReLU};
Hosts: \textsf{linear}(22) $\to$ \textsf{linear}(10) $\to$ \textsf{linear}(8) $\to$ \textsf{linear}(4) $\to$ \textsf{linear}(1)\\
\textbf{Parkinsons.}
Guests: \textsf{linear}(94) $\to$ \textsf{linear}(47) $\to$  \textsf{ReLU}; Hosts: \textsf{linear}(94) $\to$ \textsf{linear}(47) $\to$ \textsf{LeakyReLU} $\to$ \textsf{linear}(22) $\to$ \textsf{linear}(10) $\to$ \textsf{LeakyReLU}  $\to$ \textsf{linear}(1) \\
\\
\textbf{{Guest-Host Model Splits for ResNet-like Models}}\\
For all our CV models (ResNet-18, ResNet56, ResNeXt-29 8x64), each guest owns its own CNN filter as well as half of the first fully connected layer. The remaining fully connected layers are owned by the host. 
\subsection{Nomenclature}
We will use the following terminology for the remainder of the paper 
\begin{itemize}
    \item \textbf{Aligned Data}: Refers to entity-aligned/private set intersection data. For example, in the case of two clients, each client inputs corresponding parts of the same image into their respective models.
    \item \textbf{Misaligned Data}: Refers to intentionally misaligned data-- the members and order of the ``misaligned" sample space are different for each guest. In this case, clients input parts of different images into their respective models.
\end{itemize}

\subsection{Experimental Setup}
\paragraph{Exploiting data outside the intersection.}To evaluate the effect of entity augmentation, we propose an experiment where the dataset is divided into $x\%$ entity-aligned data and $\frac{(100-x)}{2}\%$ misaligned data for two clients. That is to say, we have $x\%$ of the dataset aligned between the two guests. where corresponding parts of the data are assigned to each client. The remaining $(100-x)\%$ is shuffled and \textbf{split evenly} between the two clients, i.e. each client gets a slice from a totally non-overlapping subset of the sample space. We attempt to train a split neural network with just the aligned data and investigate the impact on performance when the misaligned data is also used via Entity Augmentation.

\paragraph{Entity Alignment vs Misaligned Augmentation.} 
To test the hypothesis that training on misaligned data can outperform aligned data given long-range semantic correlations, we conduct experiments on fully aligned and intentionally misaligned data. For each dataset, we train models on both aligned and misaligned data. We compare the performance of the models to assess if misaligned data with sufficient long-range semantic correlations can lead to better learning outcomes. The results of these experiments demonstrate the impact of data alignment on model performance and the improved performance of entity augmentation.

\subsection{Implementation Details}
For the CV datasets, we apply the proposed algorithm using ResNet and ResNeXt architectures. For tabular and multiview datasets, we employ the SplitNN architecture. Each experiment is run for 60 epochs, with two guests for the CV and tabular datasets. For multiview datasets, we set the number of guests to be equal to the number of views. We implement our models in PyTorch and train them to minimize binary cross entropy loss. The PyTorch implementation internally calculates a sigmoid. We use the Adam optimizer with $\beta_1=0.9, \beta_2=0.999$. We use a learning rate of $0.001$ for all CV experiments, $0.1$ for both multiview datasets, and $5\times10^{-4}$ for both tabular datasets. 

\begin{table}[]
\centering
\caption{We compare our method to the results on vanilla VFL and 5-shot VFL due to Sun et al. \cite{Sun_2023_ICCV} using the same model. We measure accuracy when a certain number of training samples (denoted in the table as $|\cap\mathcal S|$) overlap between the guests, and the remaining samples are split evenly between the two. The host is assumed to know labels for all samples. Entity Augmentation cannot exploit as many samples as few-shot VFL but significantly more than standard VFL, and as a result is performant without requiring guests to guess private labels.}
\label{tab:fewshot}
\begin{tabular}{@{}ccccccc@{}}
\toprule
\multirow{2}{*}{\textsc{Method}} & \multirow{2}{*}{\textsc{Samples Used}} & \multirow{2}{*}{\textsc{Privacy}} & \multicolumn{4}{c}{\textsc{Accuracy (\%) at $\left|\displaystyle\cap\mathcal S\right|$}} \\ \cmidrule(l){4-7} 
                        &                                     &                                     & 256         & 512         & 1024         & 2048         \\ \midrule
Entity Aug. (Ours)             & $\displaystyle\bigcup_{i=1}^{|\mathcal G|} \mathcal S_i\cap S_h$                                  & \checkmark                                 & 67.82         & 73.35         & 73.87          & 74.25          \\
Few-shot VFL \cite{Sun_2023_ICCV}            & $\displaystyle\bigcup_{i=1}^{|\mathcal G|} \mathcal S_i$                                & \text{\sffamily X}                                           & 78.93           & 83.03            & 85.68 &   87.23          \\
Standard VFL \cite{Sun_2023_ICCV}            & $\mathcal S_h\cap\left(\displaystyle\bigcap_{i=1}^{|\mathcal G|} \mathcal S_i\right)$                                & \checkmark                                 & 31.47          & 35.33           & 42.71            & 50.75            \\\bottomrule
\end{tabular}
\end{table}

\section{Results and Discussions}
\begin{table}
\centering
\caption{Accuracy comparison between entity aligned and entity misaligned data with Entity Augmentation on MNIST and CIFAR datasets.}
\begin{tabular}{lccc}
\hline
\multirow{3}{*}{\textsc{Dataset}} & & \multicolumn{2}{c}{\textsc{Accuracy}} \\
\cline{3-4}
& \textsc{Architecture} & \textit{Aligned} & \textit{Misaligned} \\
\hline
\multirow{3}{*}{CIFAR10}& ResNet-18 & 72.92\% & 74.34\% \\
& ResNet-56 & 77.09\% & 79.12\% \\
& ResNeXt-29 8x64 & 81.08\% & 82.06\%\\
\hline
\multirow{3}{*}{MNIST}& ResNet-18 & 99.25\% & 98.20\% \\
& ResNet-56 & 99.34\% & 98.44\% \\
& ResNeXt-29 8x64 & 99.12\% & 98.43\%\\
\hline
\end{tabular}
\label{tab:mnist_cifar}
\end{table}

\paragraph{Entity Alignment vs Misaligned Augmentation.}
Our experiments with entity augmentation, as shown in Tables \ref{tab:mnist_cifar} and \ref{table2}, demonstrate that our method achieves comparable results on the MNIST dataset and improved performance on the CIFAR, Handwritten, Caltech-7, Credit Card and Parkinson's datasets. This is not unexpected since Entity Augmentation is functionally a form of CutMix, which has been shown to have a regularizing effect. \cite{yun2019cutmix}
 
MNIST, with its single color channel and simpler, well-defined shapes, presents fewer long-range feature variations compared to datasets with complex imagery. For instance, a straight line in the top quarter could ambiguously belong to a 5 or 7. Thus, performance gains from CutMix are less pronounced on MNIST.

\begin{figure*}[ht]
\centering
\caption{Training Curves for CIFAR, MNIST, Handwritten, Caltech-7, Parkinson's, and Credit Card datasets. Significant convergence improvements are observed with our method on CIFAR and MNIST. The efficacy extends to datasets like Handwritten, Caltech-7, Credit Card, and Parkinson's.}
\label{fig:combined_curves}
\begin{subfigure}{.19\textwidth}
  \centering
  \includegraphics[width=\linewidth]{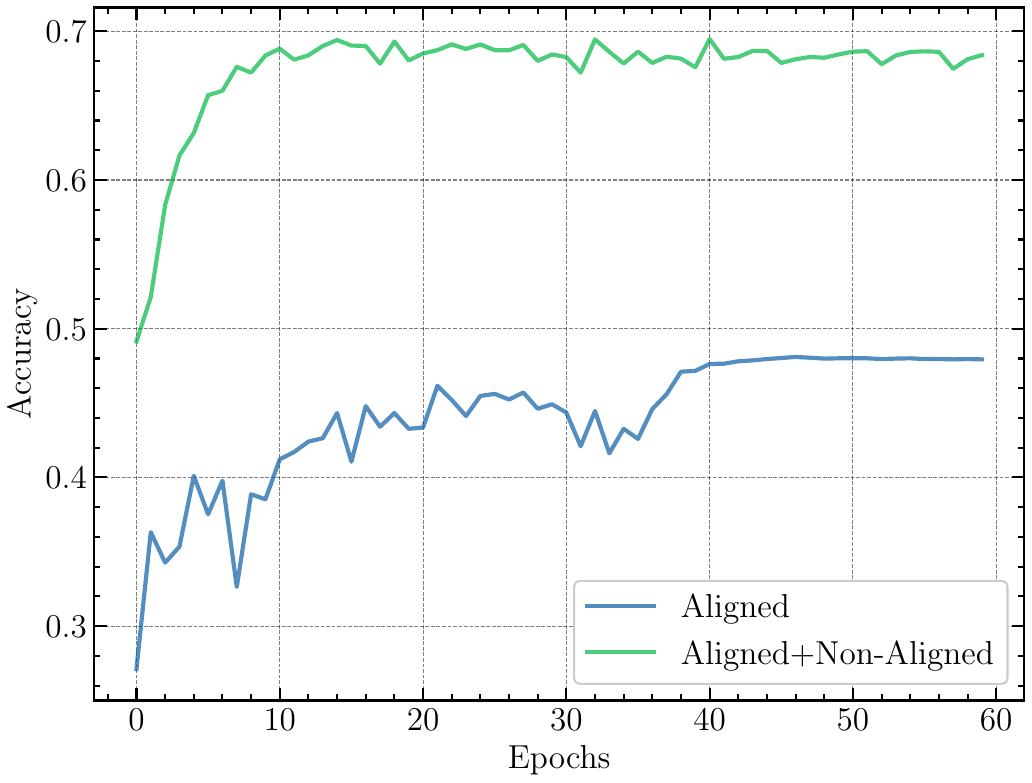}
  \caption*{CIFAR 5\%}
\end{subfigure}%
\begin{subfigure}{.19\textwidth}
  \centering
  \includegraphics[width=\linewidth]{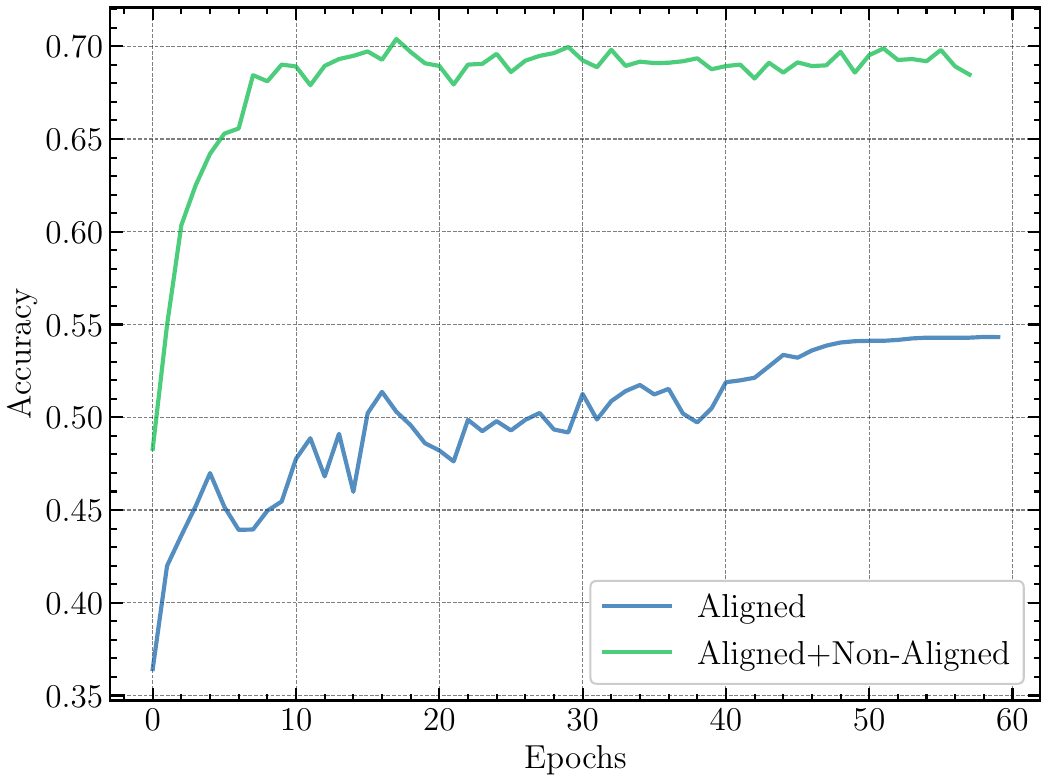}
  \caption*{CIFAR 10\%}
\end{subfigure}
\begin{subfigure}{.19\textwidth}
  \centering
  \includegraphics[width=\linewidth]{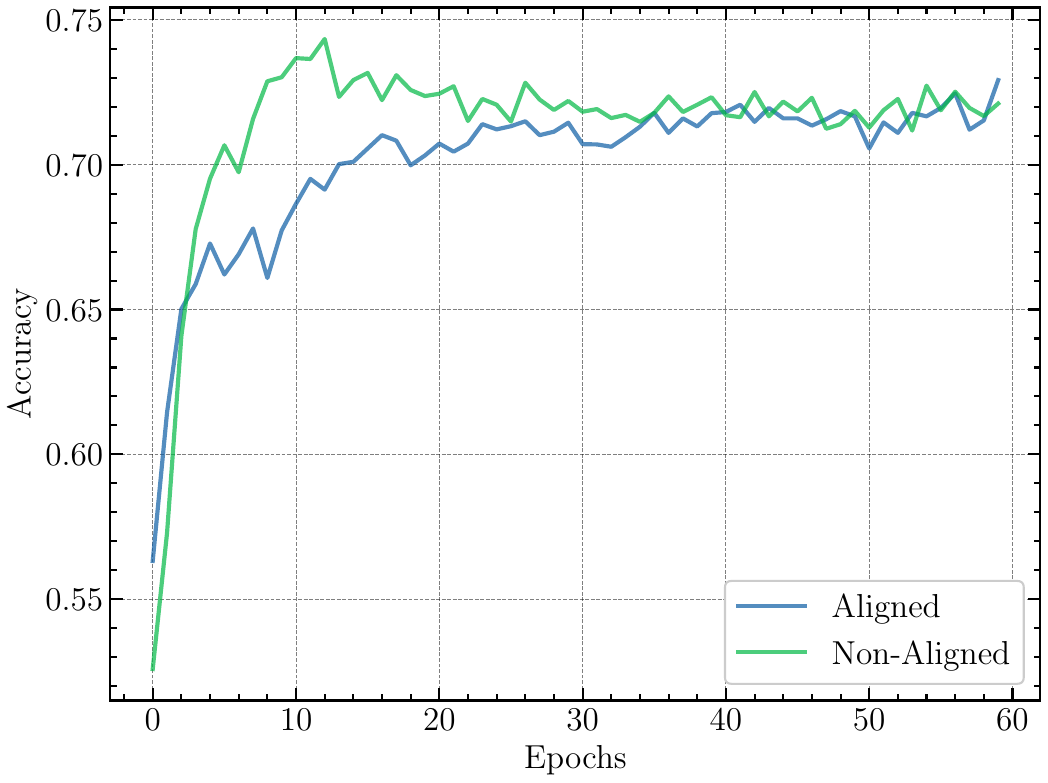}
  \caption*{CIFAR ResNet-18}
\end{subfigure}
\begin{subfigure}{.19\textwidth}
  \centering
  \includegraphics[width=\linewidth]{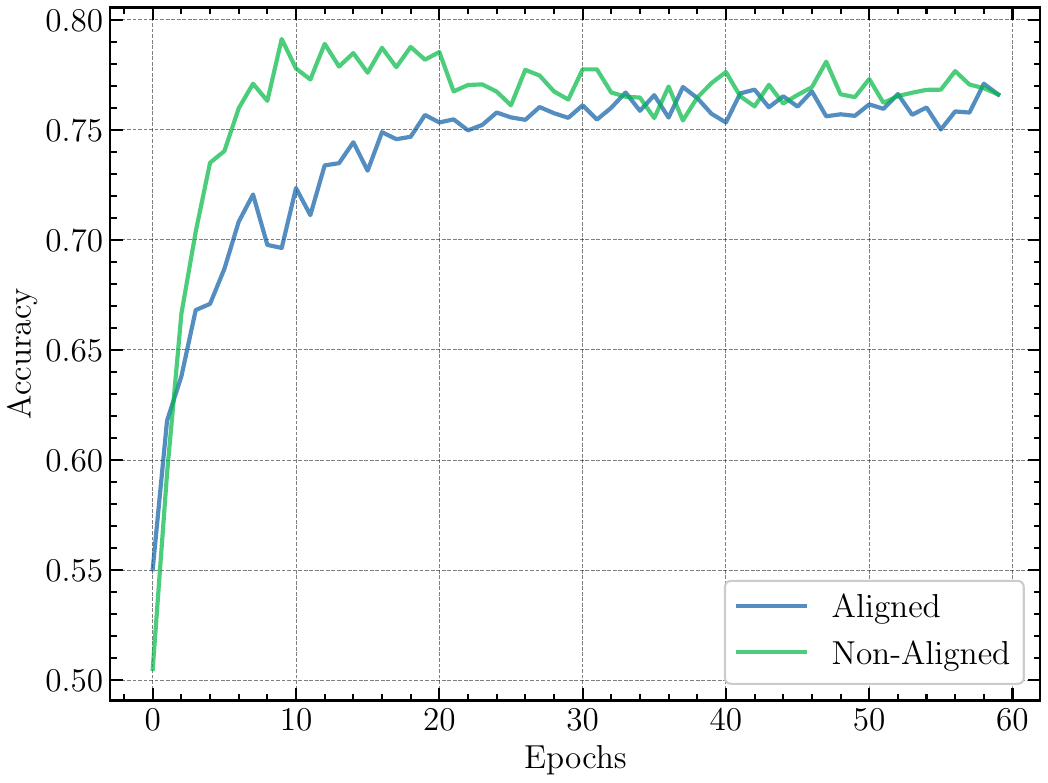}
  \caption*{CIFAR ResNet-56}
\end{subfigure}
\begin{subfigure}{.19\textwidth}
  \centering
  \includegraphics[width=\linewidth]{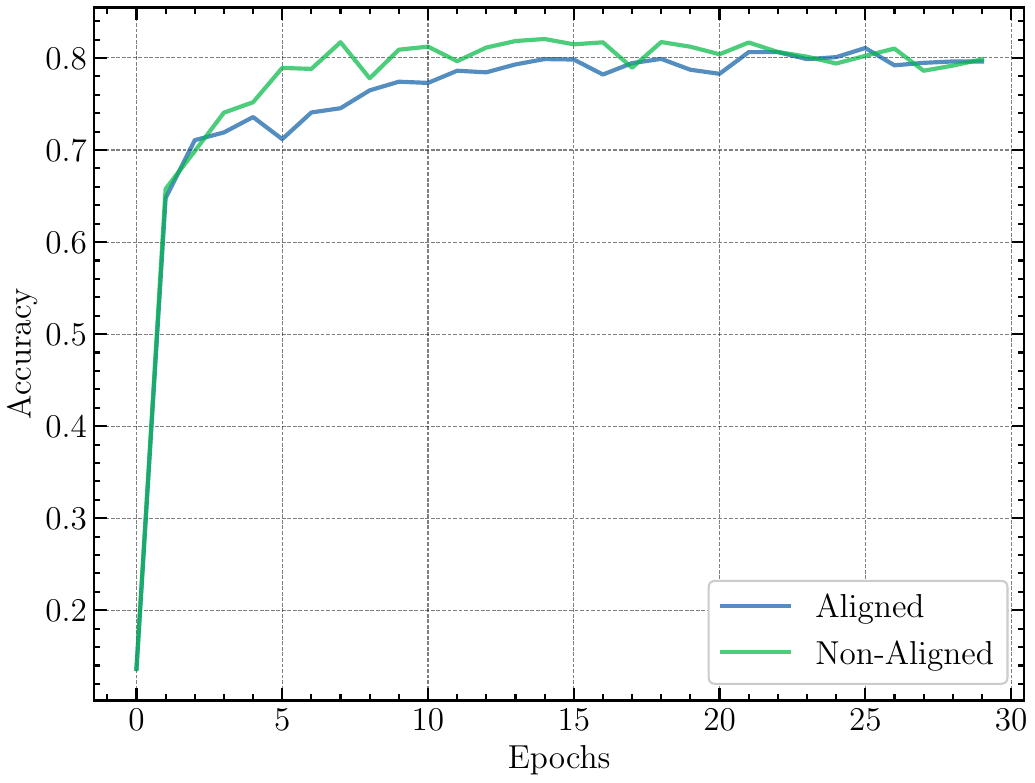}
  \caption*{CIFAR ResNext}
\end{subfigure}

\begin{subfigure}{.19\textwidth}
  \centering
  \includegraphics[width=\linewidth]{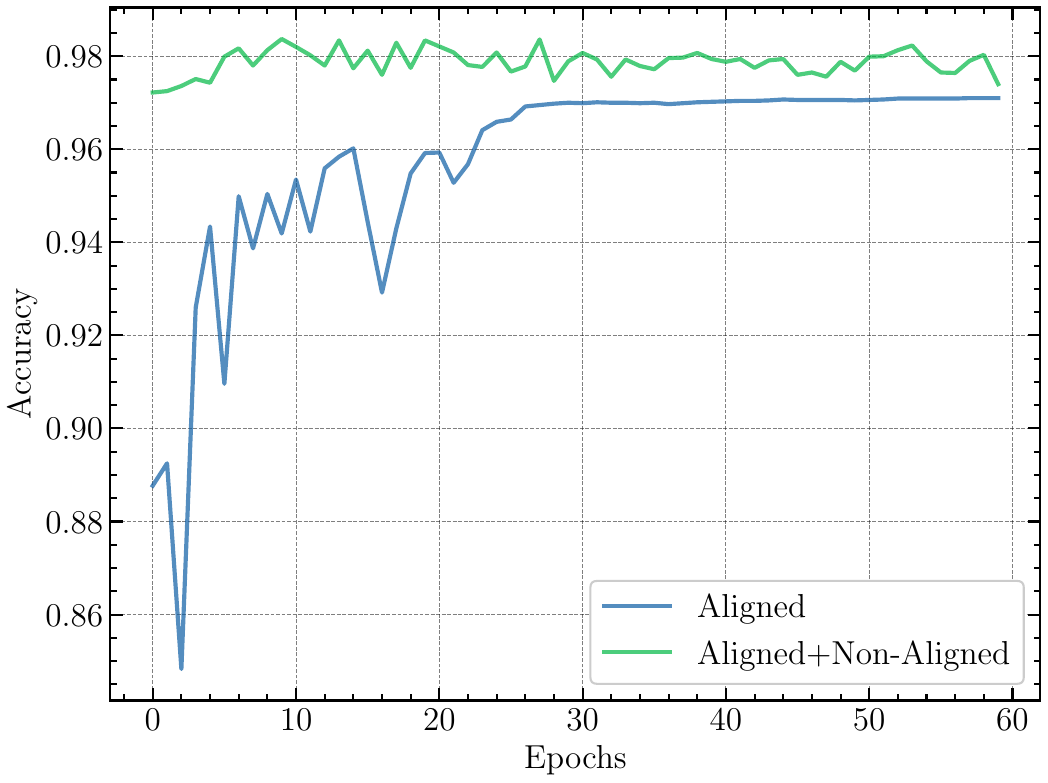}
  \caption*{MNIST 5\%}
\end{subfigure}%
\begin{subfigure}{.19\textwidth}
  \centering
  \includegraphics[width=\linewidth]{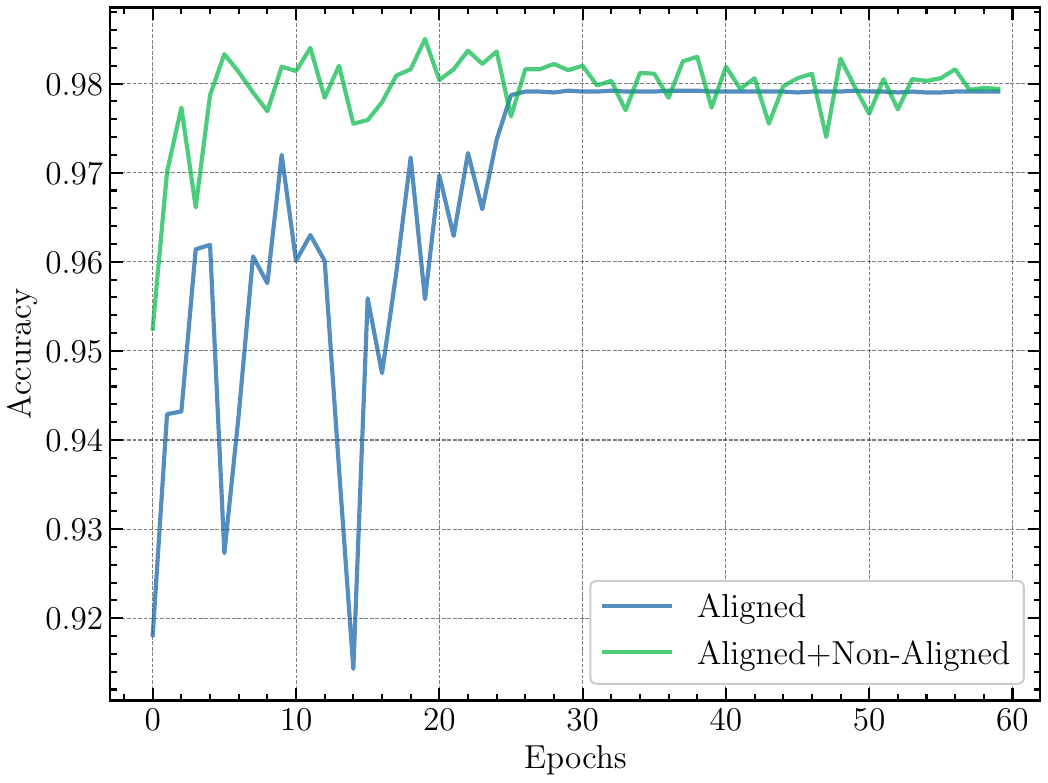}
  \caption*{MNIST 10\%}
\end{subfigure}
\begin{subfigure}{.19\textwidth}
  \centering
  \includegraphics[width=\linewidth]{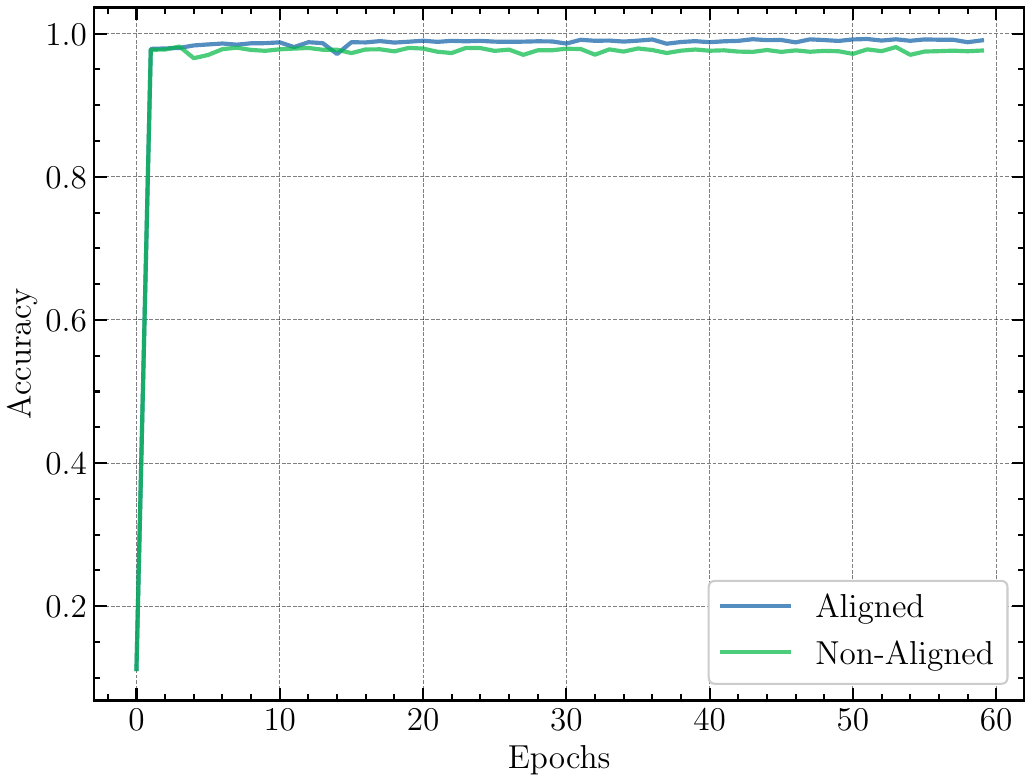}
  \caption*{MNIST ResNet-18}
\end{subfigure}
\begin{subfigure}{.19\textwidth}
  \includegraphics[width=\linewidth]{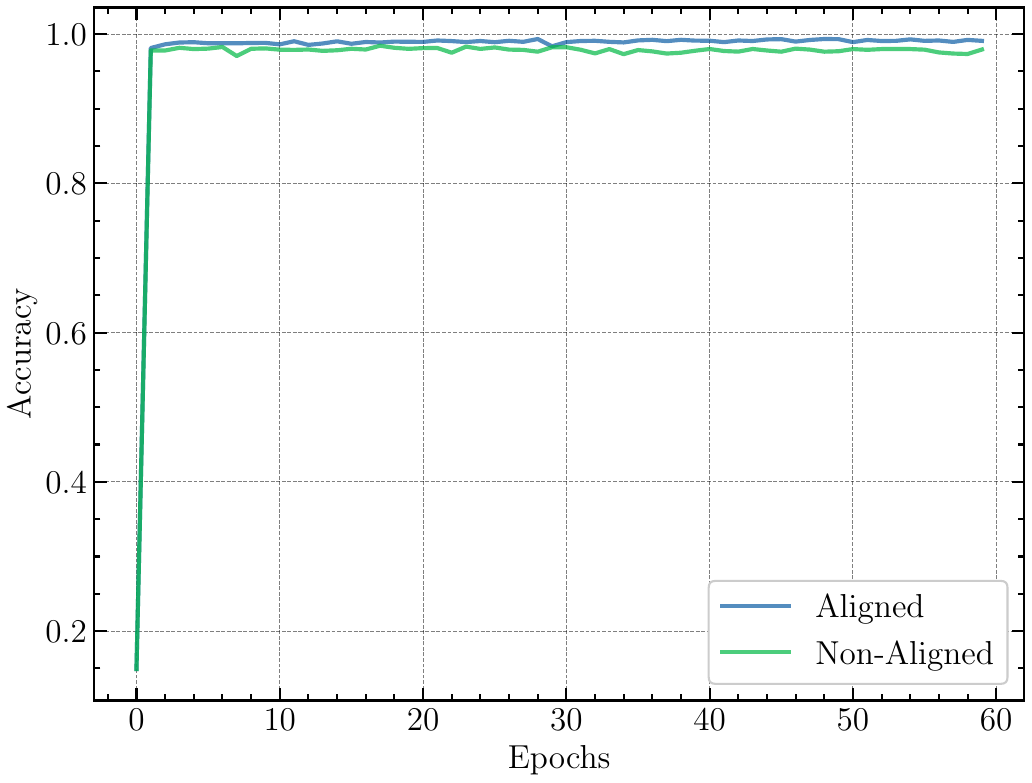}
  \caption*{MNIST ResNet-56}
\end{subfigure}
\begin{subfigure}{.19\textwidth}
  \includegraphics[width=\linewidth]{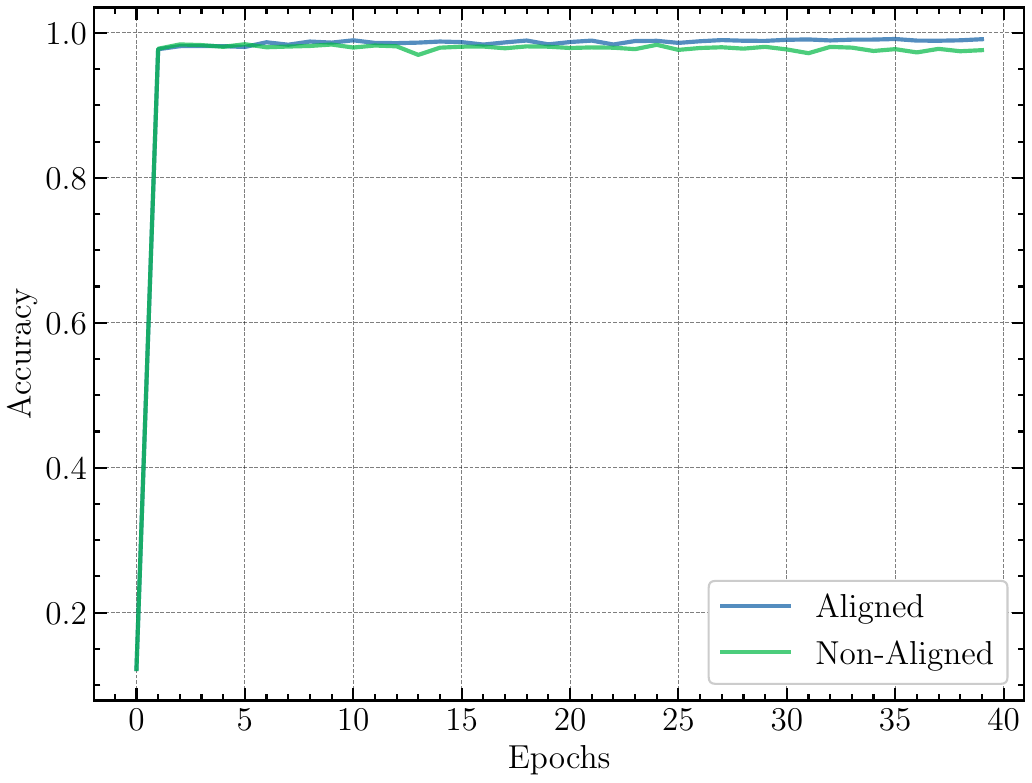}
  \caption*{MNIST ResNext}
\end{subfigure}

\begin{subfigure}{.18\textwidth}
  \centering
  \includegraphics[width=\linewidth]{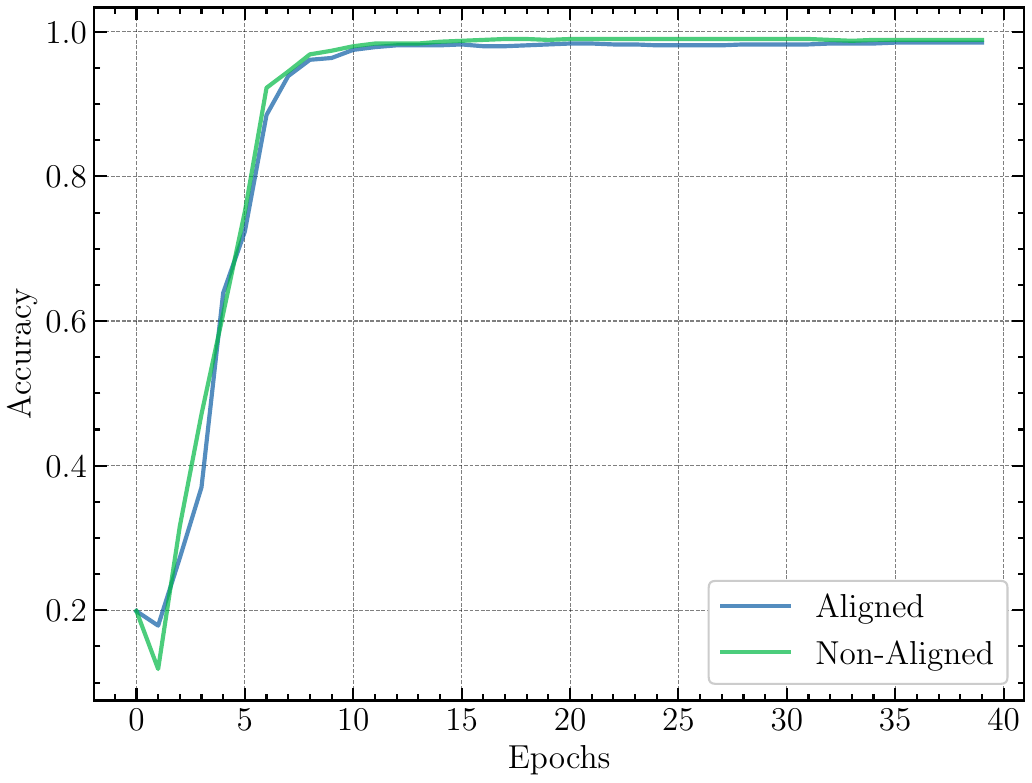}
  \caption*{Handwritten}
\end{subfigure}%
\begin{subfigure}{.18\textwidth}
  \includegraphics[width=\linewidth]{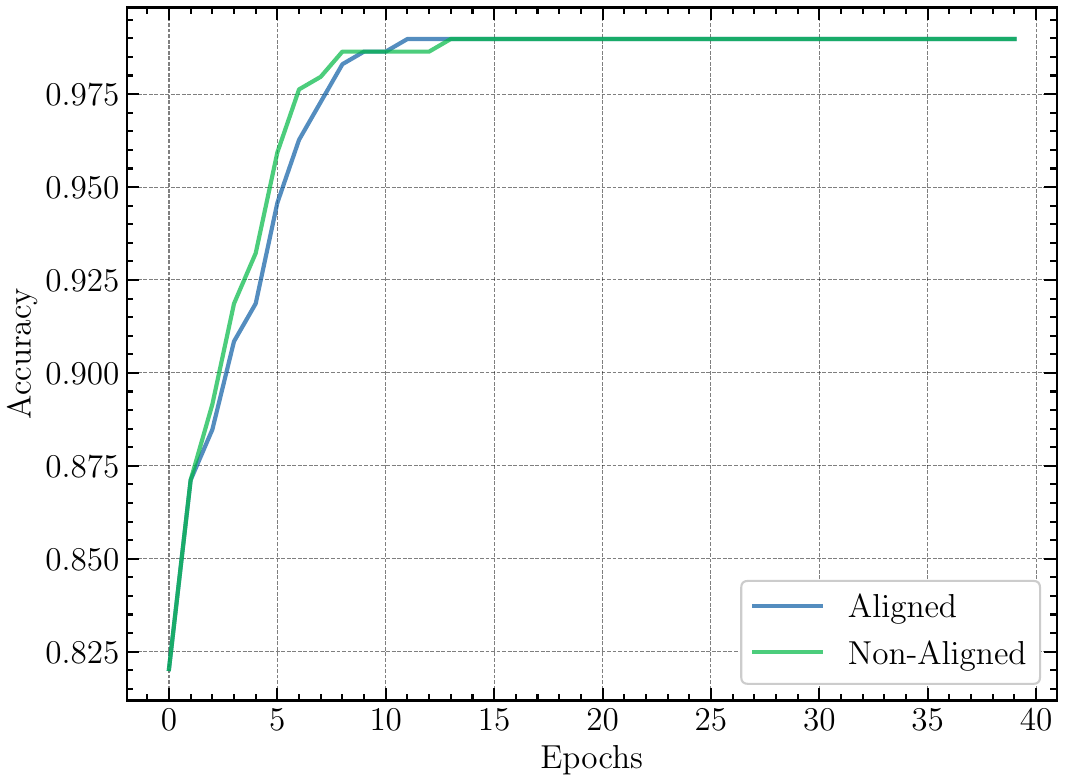}
  \caption*{Caltech-7}
\end{subfigure}
\begin{subfigure}{.18\textwidth}
  \includegraphics[width=\linewidth]{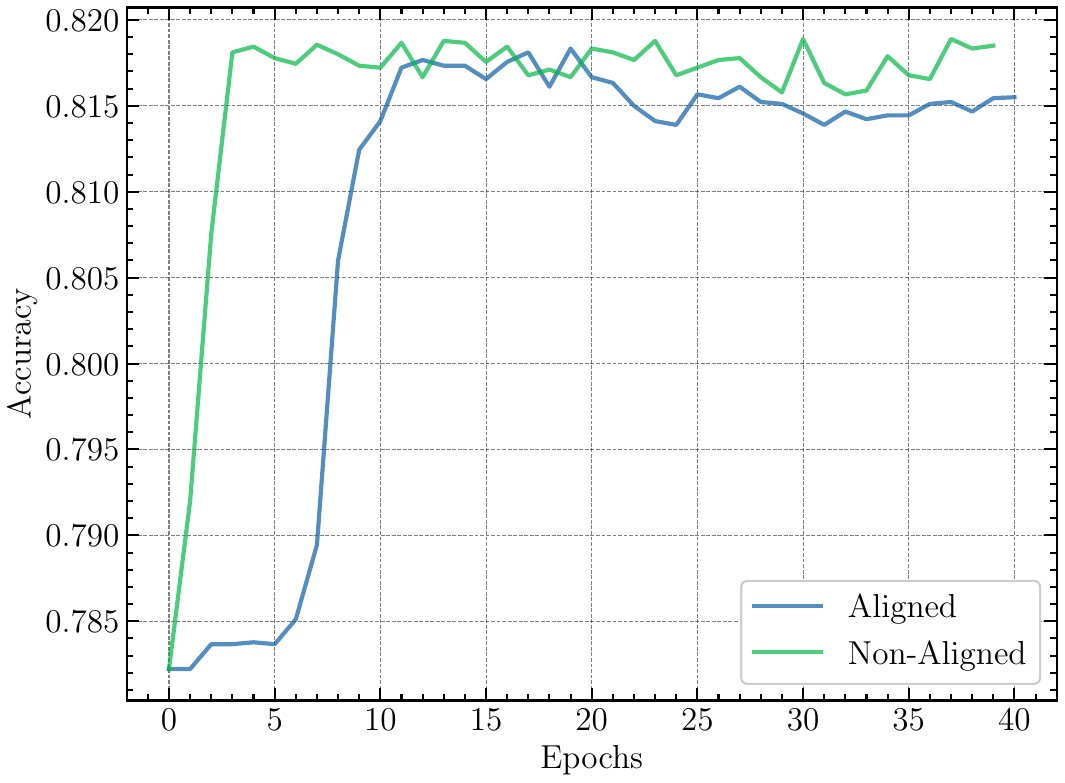}
  \caption*{Credit Card}
\end{subfigure}
\begin{subfigure}{.18\textwidth}
  \includegraphics[width=\linewidth]{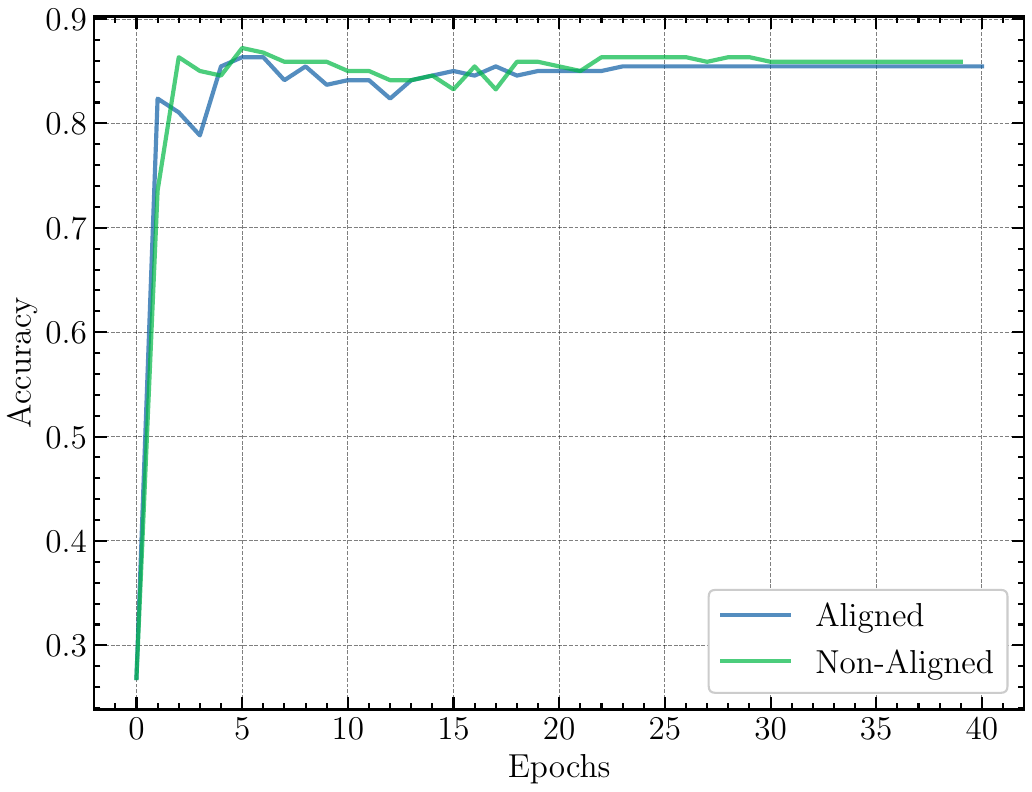}
  \caption*{Parkinson's}
\end{subfigure}
\label{fig:all_curves}
\end{figure*}

\begin{table}
\centering
\caption{Accuracy comparison between training on entity-aligned data vs entity augmented misaligned data on the view-partitioned datasets Handwritten and Caltech-7, and the vertically partitioned tabular datasets Credit Card and Parkinsons. }
\begin{tabular}{lcc}
\hline
\textsc{Dataset} & \textsc{Aligned} & \textsc{Misaligned} \\
\hline
Handwritten & 98.25\% & 98.63\% \\
Caltech-7 & 98.98\% & 98.98\% \\
Credit Card & 81.55\% & 81.95\%\\
Parkinsons & 86.34\% & 87.22\% \\
\hline
\end{tabular}
\label{table2}
\end{table}
\paragraph{Exploiting data outside the intersection.}
From the results of our experiment in Table 3, it is visible that when only a tiny entity-aligned dataset is available, using entity misaligned/augmented data (i.e., with no private set intersection) along with it for training provides better performance compared to training only on the aligned dataset. These results clearly support our claim that entity-misaligned/augmented data is helpful for training and results in better performance than only using entity-aligned data, resulting in seamless integration of diverse data sources, reduced data wastage, and enhanced model learning efficiency.
\begin{table}
\centering
\caption{Accuracy comparison between training a ResNet--18 on only a small intersection of entity-aligned data vs that entity-aligned data combined with entity augmented misaligned data on MNIST and CIFAR datasets. }
\begin{tabular}{lcc}
\hline
\multirow{1}{*}{\textsc{Dataset}} &  \multicolumn{2}{c}{\textsc{Accuracy}} \\
\cline{2-3}
& \textit{Aligned} & \textit{Aligned + Misaligned} \\
\hline
\multirow{1}{*}{CIFAR10 ($x=5\%$)}& 48.1\% & 69.48\%  \\
\multirow{1}{*}{MNIST ($x=5\%$)} & 97.1\% & 98.37\%  \\
\multirow{1}{*}{CIFAR10 ($x=10\%$)} & 54.34\% & 70.4\%  \\
\multirow{1}{*}{MNIST ($x=10\%$)} & 97.92\% & 98.5\%  \\
\hline
\end{tabular}
\label{tab:table3}
\end{table}

\paragraph{More efficient training.} 
Figures \ref{fig:combined_curves} reveal that Entity Augmentation not only boosts the skyline performance of VFL models, but also allow them to converge substantially faster. Experiments using only $x\%$ aligned data plateaus at a much lower accuracy \textit{and} at a far earlier epoch. A similar trend may be seen in our experiments on fully aligned vs fully misaligned data. Another interesting phenomenon is the stability of training-- the test accuracy is qualitatively smoother and more stable wherever Entity Augmentation is used. 

\section{Future Work}
The proposed method shows promising results for training pipelines where the label can be represented in a one-hot encoded fashion. Subsequently, we seek to extend the idea of generating synthetic labels for regressive tasks. In this light, Verma et al. \cite{verma2019manifold} investigate the potential of swapping weights in the penultimate layer to create samples through inference. Expanding upon this, Hwang et al. \cite{hwang2022regmix} use linear interpolation and constrained sampling for data augmentation. Furthermore, Jiang et al. \cite{10166693} employ Gaussian Mixture Models to facilitate the generation of synthetic and continuous sensor data. Our future endeavours will focus on incorporating such augmentation techniques within the Vertical Federated Learning (VFL) framework. This integration seeks to optimize the utilization of data that lies beyond the confines of the Private Set Intersection, thereby enhancing the efficiency and effectiveness of the VFL pipeline for regressive tasks.

\section{Conclusion}
This work presents Entity Augmentation, a strategy for generating semantically meaningful labels for guest activations without entity alignment. We interpolate labels weighted by features to synthesize labels for training. We subsequently demonstrate that our pipeline achieves performance on par with traditional FL approaches that require entity alignment. Our evaluations on the CIFAR10 and MNIST datasets showed improved results across various baseline architectures, and we achieved competitive results on Handwritten, Caltech-7, Parkinsons and Credit Card datasets. In future, we seek to extend the augmentation technique to regressive tasks and experiment with Gaussian mixture models and constrained sampling.

%
%
%
\bibliographystyle{splncs04}
\bibliography{main}

\begin{thebibliography}{10}
\providecommand{\url}[1]{\texttt{#1}}
\providecommand{\urlprefix}{URL }
\providecommand{\doi}[1]{https://doi.org/#1}

\bibitem{amalanshu2024decoupled}
Amalanshu, A., Sirvi, Y., Inouye, D.I.: Decoupled vertical federated learning for practical training on vertically partitioned data (2024). arXiv:2403.03871

\bibitem{ceballos2020splitnndriven}
Ceballos, I., Sharma, V., Mugica, E., Singh, A., Roman, A., Vepakomma, P., Raskar, R.: Splitnn-driven vertical partitioning (2020). arXiv:2008.04137

\bibitem{Dua:2019}
Dua, D., Graff, C.: {UCI} machine learning repository (2017), \url{http://archive.ics.uci.edu/ml}

\bibitem{Gupta2018}
Gupta, O., Raskar, R.: Distributed learning of deep neural network over multiple agents. Journal of Network and Computer Applications  \textbf{116}, ~1--8 (Aug 2018). \doi{10.1016/j.jnca.2018.05.003}, \url{https://doi.org/10.1016/j.jnca.2018.05.003}

\bibitem{hardy2017private}
Hardy, S., Henecka, W., Ivey-Law, H., Nock, R., Patrini, G., Smith, G., Thorne, B.: Private federated learning on vertically partitioned data via entity resolution and additively homomorphic encryption (2017). arXiv:1711.10677

\bibitem{he2015deep}
He, K., Zhang, X., Ren, S., Sun, J.: Deep residual learning for image recognition (2015). arXiv:1512.03385

\bibitem{hwang2022regmix}
Hwang, S.H., Whang, S.E.: Regmix: Data mixing augmentation for regression (2022). arXiv:2106.03374

\bibitem{10166693}
Jiang, X., Yao, L., Yang, Z., Song, Z., Shen, B.: Gaussian mixture model and double-weighted deep neural networks for data augmentation soft sensing. In: 2023 IEEE 12th Data Driven Control and Learning Systems Conference (DDCLS). pp. 1914--1919 (2023). \doi{10.1109/DDCLS58216.2023.10166693}

\bibitem{Krizhevsky09learningmultiple}
Krizhevsky, A.: Learning multiple layers of features from tiny images. Tech. rep. (2009)

\bibitem{lecun2010mnist}
LeCun, Y., Cortes, C., Burges, C.: Mnist handwritten digit database. ATT Labs [Online]. Available: http://yann.lecun.com/exdb/mnist  \textbf{2} (2010)

\bibitem{li_andreeto_ranzato_perona_2022}
Li, F.F., Andreeto, M., Ranzato, M., Perona, P.: Caltech 101 (April 2022). \doi{10.22002/D1.20086}

\bibitem{Lu2020MultiPartyPS}
Lu, L., Ding, N.: Multi-party private set intersection in vertical federated learning. 2020 IEEE 19th International Conference on Trust, Security and Privacy in Computing and Communications (TrustCom) pp. 707--714 (2020), \url{https://api.semanticscholar.org/CorpusID:231916141}

\bibitem{pmlr-v54-mcmahan17a}
McMahan, B., Moore, E., Ramage, D., Hampson, S., Arcas, B.A.y.: {Communication-Efficient Learning of Deep Networks from Decentralized Data}. In: Singh, A., Zhu, J. (eds.) Proceedings of the 20th International Conference on Artificial Intelligence and Statistics. Proceedings of Machine Learning Research, vol.~54, pp. 1273--1282. PMLR (20--22 Apr 2017), \url{https://proceedings.mlr.press/v54/mcmahan17a.html}

\bibitem{MORALES2023100567}
Morales, D., Agudo, I., Lopez, J.: Private set intersection: A systematic literature review. Computer Science Review  \textbf{49},  100567 (2023). \doi{https://doi.org/10.1016/j.cosrev.2023.100567}, \url{https://www.sciencedirect.com/science/article/pii/S1574013723000345}

\bibitem{nock2018entity}
Nock, R., Hardy, S., Henecka, W., Ivey-Law, H., Patrini, G., Smith, G., Thorne, B.: Entity resolution and federated learning get a federated resolution (2018). arXiv:1803.04035

\bibitem{sakar2019comparative}
Sakar, C.O., Serbes, G., Gunduz, A., Tunc, H.C., Nizam, H., Sakar, B.E., Tutuncu, M., Aydin, T., Isenkul, M.E., Apaydin, H.: A comparative analysis of speech signal processing algorithms for parkinson’s disease classification and the use of the tunable q-factor wavelet transform. Applied Soft Computing  \textbf{74},  255--263 (2019)

\bibitem{Sun_2023_ICCV}
Sun, J., Xu, Z., Yang, D., Nath, V., Li, W., Zhao, C., Xu, D., Chen, Y., Roth, H.R.: Communication-efficient vertical federated learning with limited overlapping samples. In: Proceedings of the IEEE/CVF International Conference on Computer Vision (ICCV). pp. 5203--5212 (October 2023)

\bibitem{verma2019manifold}
Verma, V., Lamb, A., Beckham, C., Najafi, A., Mitliagkas, I., Courville, A., Lopez-Paz, D., Bengio, Y.: Manifold mixup: Better representations by interpolating hidden states (2019). arXiv:1806.05236

\bibitem{xie2017aggregated}
Xie, S., Girshick, R., Dollár, P., Tu, Z., He, K.: Aggregated residual transformations for deep neural networks (2017). arXiv:1611.05431

\bibitem{yeh2009comparisons}
Yeh, I.C., Lien, C.h.: The comparisons of data mining techniques for the predictive accuracy of probability of default of credit card clients. Expert systems with applications  \textbf{36}(2),  2473--2480 (2009)

\bibitem{yun2019cutmix}
Yun, S., Han, D., Oh, S.J., Chun, S., Choe, J., Yoo, Y.: Cutmix: Regularization strategy to train strong classifiers with localizable features (2019). arXiv:1905.04899

\end{thebibliography}

%




\end{document}